\documentclass{llncs}

\usepackage{amsfonts}
\usepackage{amsmath}
\usepackage{arydshln}  % dash line in table
\usepackage[table]{xcolor}  % for color in table
\usepackage{graphics}  % for scalebox
\usepackage{multirow}  % for multirow in table 
\def\b{\ensuremath\boldsymbol}
\usepackage{graphicx}

\usepackage{amssymb}

\usepackage{url}
\usepackage[algo2e,linesnumbered,lined,boxed,vlined]{algorithm2e}

\usepackage{tikz}
\usepackage{lipsum}
\newcommand\copyrighttextt{%
  \footnotesize Accepted (to appear) in International Conference on Image Analysis and Recognition (ICIAR) 2020, Springer.}
\newcommand\copyrightnotice{%
\begin{tikzpicture}[remember picture,overlay]
\node[anchor=south,yshift=10pt] at (current page.south) {\fbox{\parbox{\dimexpr\textwidth-\fboxsep-\fboxrule\relax}{\copyrighttextt}}};
\end{tikzpicture}%
}

\begin{document}

\title{Theoretical Insights into the Use of Structural Similarity Index In Generative Models and Inferential Autoencoders}

\author{Benyamin Ghojogh,
Fakhri Karray,
Mark Crowley
}
\authorrunning{F. Author et al.}
\institute{Department of Electrical and Computer Engineering, \\ University of Waterloo, Waterloo, ON, Canada  \\
\email{\{bghojogh, karray, mcrowley\}@uwaterloo.ca} 
}

% \author{Benyamin Ghojogh,
% Milad Sikaroudi,
% Hamid R. Tizhoosh,
% Fakhri Karray,
% Mark Crowley
% }

% %
% \institute{Department of Electrical and Computer Engineering,\\ University of Waterloo, Waterloo, ON, Canada  \\
% \email{\{bghojogh, karray, mcrowley\}@uwaterloo.ca}
% }

\maketitle              % typeset the title of the contribution

\begin{abstract}
Generative models and inferential autoencoders mostly make use of $\ell_2$ norm in their optimization objectives. 
In order to generate perceptually better images, this short paper theoretically discusses how to use Structural Similarity Index (SSIM) in generative models and inferential autoencoders. We first review SSIM, SSIM distance metrics, and SSIM kernel. We show that the SSIM kernel is a universal kernel and thus can be used in unconditional and conditional generated moment matching networks. Then, we explain how to use SSIM distance in variational and adversarial autoencoders and unconditional and conditional Generative Adversarial Networks (GANs). Finally, we propose to use SSIM distance rather than $\ell_2$ norm in least squares GAN. 
\keywords{generative moment matching network, generative adversarial network, variational autoencoder, adversarial autoencoder, structural similarity index, SSIM kernel, perceptual image generation}
\end{abstract}

\copyrightnotice

\section{Introduction}

Learning models can be divided into discriminative and generative \cite{ng2002discriminative}. Many of the generative models and inferential autoencoders produce blurry images for different reasons. Variational Autoencoder (VAE) \cite{doersch2016tutorial} has this flaw maybe because of the lower bound approximation or restriction on the distribution. However, another reason might be the use of a non-perceptual distance in its objective \cite{wang2009mean}. 
Unconditional and conditional Generative Moment Matching Networks (GMMNs) \cite{li2015generative,ren2016conditional} also use radial basis function kernel having $\ell_2$ norm. Adversarial Autoencoder (AAE) \cite{makhzani2015adversarial} and unconditional/conditional Generative Adversarial Networks (GANs) \cite{goodfellow2014generative,mirza2014conditional} also use non-perceptual metrics in their objectives for comparison of real and fake data. Least Squares GAN (LSGAN) \cite{mao2017least} uses $\ell_2$ norm or Mean Square Error (MSE) in its loss function.
However, MSE is shown not to be perfect for image quality assessment \cite{wang2009mean}. 
Structural Similarity Index (SSIM) \cite{wang2004image} is a perceptual measure for image quality. In this paper, we theoretically explain how SSIM can be used in different generative models and inferential autoencoders. 
Using SSIM can \textit{improve the perceptual quality of the generated images} by these models. 
This is a poster paper and according to the expectation of the conference from a short poster paper, we suffice to the theoretical analysis and defer the empirical results to future work.

\section{Structural Similarity Index, Image Structure Subspace, and SSIM Kernel}\label{section_SSIM}

Consider two reshaped images $\b{x}_i, \b{x}_j \in \mathbb{R}^d$. 
The SSIM between two reshaped image blocks $\breve{\b{x}}_i = [x_i^{(1)}, \dots, x_i^{(q)}]^\top \in \mathbb{R}^q$ and $\breve{\b{x}}_j = [x_j^{(1)}, \dots, x_j^{(q)}]^\top \in \mathbb{R}^q$, in color intensity range $[0,l]$, is: $\mathbb{R} \ni \text{SSIM}(\breve{\b{x}}_i, \breve{\b{x}}_j) := [(2 \mu_{x_i} \mu_{x_j} + c_1)/(\mu_{x_i}^2 + \mu_{x_j}^2 + c_1)] [(2 \sigma_{x_i} \sigma_{x_j} + c_2)/(\sigma_{x_i}^2 + \sigma_{x_j}^2 + c_2)] [(\sigma_{x_i, x_j} + c_3)/(\sigma_{x_i}\sigma_{x_j} + c_3)]$, where $\mu_{x_i} = (1/q) \sum_{k=1}^q x_i^{(k)}$, $\sigma_{x_i} = \Big[\big(1/(q-1)\big) \sum_{k=1}^q (x_i^{(k)} - \mu_{x_i})^2\Big]^{0.5}$, $\sigma_{x_i,x_j} = \big(1/(q-1)\big) \sum_{k=1}^q (x_i^{(k)} - \mu_{x_i}) (x_j^{(k)} - \mu_{x_j})$, $c_1=(0.01 \times l)^2$, $c_2=2\,c_3=(0.03 \times l)^2$, and $\mu_{x_j}$ and $\sigma_{x_j}$ are defined similarly for $\breve{\b{x}}_j$ \cite{wang2004image}.
Since $c_2=2\,c_3$, we can simplify SSIM to $\text{SSIM}(\breve{\b{x}}_i, \breve{\b{x}}_j) = s_1(\breve{\b{x}}_i, \breve{\b{x}}_j) \times s_2(\breve{\b{x}}_i, \breve{\b{x}}_j)$, where $s_1(\breve{\b{x}}_i, \breve{\b{x}}_j) := (2 \mu_{x_i} \mu_{x_j} + c_1)/(\mu_{x_i}^2 + \mu_{x_j}^2 + c_1)$ and $s_2(\breve{\b{x}}_i, \breve{\b{x}}_j) := (2 \sigma_{x_i, x_j} + c_2)/(\sigma_{x_i}^2 + \sigma_{x_j}^2 + c_2)$.
If the vectors $\breve{\b{x}}_i$ and $\breve{\b{x}}_j$ have zero mean, i.e., $\mu_{x_i} = \mu_{x_j} = 0$, the SSIM becomes $\mathbb{R} \ni \text{SSIM}(\breve{\b{x}}_i, \breve{\b{x}}_j) = (2\breve{\b{x}}_i^\top \breve{\b{x}}_j + c) / (||\breve{\b{x}}_i||_2^2 + ||\breve{\b{x}}_j||_2^2 + c)$, where $c = (q-1) \,c_2$ \cite{otero2014unconstrained}.
The distance based on SSIM, which we denote by $||.||_S$, is \cite{otero2014unconstrained,brunet2012mathematical}:
\begin{align}\label{equation_SSIM_distance}
\mathbb{R} \ni ||\breve{\b{x}}_i - \breve{\b{x}}_j||_S := \sqrt{1 -  \text{SSIM}(\breve{\b{x}}_i, \breve{\b{x}}_j)} = \bigg[\frac{||\breve{\b{x}}_i - \breve{\b{x}}_j||_2^2}{||\breve{\b{x}}_i||_2^2 + ||\breve{\b{x}}_j||_2^2 + c}\bigg]^{0.5},
\end{align}
where $\mu_{x_i} = \mu_{x_j} = 0$. Note that if the means of blocks $\breve{\b{x}}_i$ and $\breve{\b{x}}_j$ are not very different, $\sqrt{1 -  \text{SSIM}(\breve{\b{x}}_i, \breve{\b{x}}_j)}$ is still a good approximation to SSIM distance even without centering the blocks \cite{brunet2012mathematical}. Some papers use this approximation and do not center the patches (cf. \cite{zhao2016loss}). 

Some works have used SSIM in machine learning for learning the image structure subspace \cite{ghojogh2019image} which captures the intrinsic features of an image in terms of structural similarity and distortions. In \cite{ghojogh2019image}, a kernel, named SSIM kernel, is proposed which can be used in kernel methods in machine learning \cite{hofmann2008kernel}. This kernel is $\b{K} = -(1/2)\, \b{H} \b{D} \b{H}$ where $\mathbb{R}^{n \times n} \ni \b{H} = \b{I} - (1/n) \b{1}\b{1}^\top$ is the centering matrix, $\b{D} \in \mathbb{R}^{n \times n}$ is the distance matrix, and $n$ is the sample size of data. Let $\b{D}'_{i,j}$ be the distance map of two images $\b{x}_i$ and $\b{x}_j$ whose entry for every patch of these images is \cite{brunet2012mathematical}:
\begin{align}\label{equation_SSIM_distance2}
\mathbb{R} \ni ||\breve{\b{x}}_i - \breve{\b{x}}_j||_S := \sqrt{2 - s_1(\breve{\b{x}}_i, \breve{\b{x}}_j) - s_2(\breve{\b{x}}_i, \breve{\b{x}}_j)}.
\end{align}
Note that one may use Eq. (\ref{equation_SSIM_distance}) for $\b{D}'$ (and in SSIM kernel) but should center every patch while Eq. (\ref{equation_SSIM_distance2}) does not require preprocessing but may be harder to compute. 
The $(i,j)$-th element of distance matrix is $\b{D}(i, j) := ||\b{D}'_{i,j}||_F$ where $||.||_F$ is the Frobenius norm.
Furthermore, note that the SSIM distance is quasi-convex \cite{brunet2012mathematical} so it is suitable for optimization \cite{brunet2018optimizing} in different applications such as machine learning \cite{ghojogh2019image}.

\section{Generative Moment Matching Network}

Maximum Mean Discrepancy (MMD), or kernel two sample test, is a measure of difference of two distributions by comparing their moments \cite{gretton2012kernel}. Let $\mathcal{X} := \{\b{x}_i\}_{i=1}^{n_x}$ and $\mathcal{Y} := \{\b{y}_i\}_{i=1}^{n_y}$ be two samples of the distributions $P_x$ and $P_y$, respectively. 
The MMD is defined as $\text{MMD}(P_x, P_y) := \sup_{f \in \mathcal{K}}(\mathbb{E}[f(\mathcal{X})] - \mathbb{E}[f(\mathcal{Y})])$ where $\mathcal{K}$ is a class of functions. If $\mathcal{K}$ is a unit ball in a universal reproducing kernel Hilbert space $\mathcal{F}$, we have: $\text{MMD}^2(P_x, P_y) = ||\frac{1}{n_x} \sum_{i=1}^{n_x} \b{\phi}(\b{x}_i) - \frac{1}{n_y} \sum_{i=1}^{n_y} \b{\phi}(\b{y}_i)||_\mathcal{F}^2$
\begin{align}
&= \frac{1}{n_x^2} \sum_{i=1}^{n_x} \sum_{j=1}^{n_y} k(x_i, x_j) + \frac{1}{n_y^2} \sum_{i=1}^{n_y} \sum_{j=1}^{n_y} k(y_i, y_j) - \frac{2}{n_x n_y} \sum_{i=1}^{n_x} \sum_{j=1}^{n_y} k(x_i, y_j), \label{equation_MMD}
\end{align}
where $\b{\phi}(.)$ is the pulling function and $k(.,.)$ is the kernel \cite{hofmann2008kernel}. 

GMMN \cite{li2015generative} uses Eq. (\ref{equation_MMD}) as the loss for training a network where the Radial Basis Function (RBF) kernel is utilized. The GMMN is a network which accepts random uniform samples in input and tries to match the moments of network's output with the batch of training data. 
It has two versions, i.e., in data space and code space. In the latter, the output layer of GMNN is the latent space of an autoencoder which is trained beforehand. 
Conditional GMMN \cite{ren2016conditional} uses Conditional MMD (CMMD) where non-uniform weights are used in MMD.  
The CMMD is defined as $||C_{\mathcal{X}|\mathcal{Z}} - C_{\mathcal{Y}|\mathcal{Z}}||_{\mathcal{F} \otimes \mathcal{G}}^2$ where $\mathcal{Z}$ is the variable conditioned on and $\otimes$ is the tensor product (see \cite{ren2016conditional} for more details). 

In GMMN and conditional GMMN, the RBF kernel is used. SSIM kernel (see Section \ref{section_SSIM}) can be used as the kernel in these two generative models. Note that only a universal kernel can be used in MMD \cite{gretton2012kernel,li2015generative} and CMMD \cite{ren2016conditional}. Paper \cite{steinwart2001influence} has shown that according to the Stone-Weierstrass theorem \cite{de1959stone}, the universal kernels can be expanded in certain types of Taylor or Fourier series. The RBF kernel is an example. It is shown in \cite{ghojogh2019image} that the SSIM kernel can be expanded by Taylor series similar to the RBF kernel; hence, \textit{SSIM kernel is a universal kernel} and thus can be used in GMMN and conditional GMMN.

\section{Variational Autoencoder}

VAE \cite{doersch2016tutorial} can be considered as the nonlinear generalization of factor analysis \cite{harman1976modern}. Training of its encoder, with weights $\b{\psi}$, and decoder, with weights $\b{\theta}$, can be seen as the E-step and M-step in expectation maximization algorithm, respectively. 
It maximizes the Evidence Lower Bound (ELBO) of the log likelihood of data \cite{kingma2013auto}. 
The loss to be minimized in VAE is:
\begin{align}\label{equation_VAE_loss}
\mathcal{L} = -\text{KL}\big(q(\b{z}|\b{x}, \b{\psi})\, ||\, p(\b{z})\big) + \mathbb{E}_{q(\b{z}|\b{x}, \b{\psi})}\big[\log p(\b{x} | \b{z}, \b{\theta})\big],
\end{align}
where $\text{KL}(.)$ is the KL-divergence, $\b{z}$ is the latent variable, $\b{x}$ is the input or re-generated data, and $q$ and $p$ are the conditional distributions in the encoder and decoder, respectively. The first and second terms in Eq. (\ref{equation_VAE_loss}) are responsible for tuning the distribution of latent variable and better generation of data out of the latent variable, respectively. The second term, which takes care of data reconstruction, is usually replaced by the cross-entropy or $\ell_2$ norm of data and generated data. 
However, $\ell_2$ norm is not perfect for image fidelity \cite{wang2009mean}. The fact that the generated images by VAE are not perceptually satisfactory has been addressed by literature \cite{goodfellow2014generative}. 
We can use SSIM distance, i.e. Eq. (\ref{equation_SSIM_distance}) or (\ref{equation_SSIM_distance2}), for the second term to \textit{measure how perceptually good the generated images are}.

\section{Generative Adversarial Networks \& Adversarial Autoencoder}

As mentioned, GAN \cite{goodfellow2014generative} is proposed to cover the perceptual lack of VAE. It claims that the $\ell_2$ norm used in VAE is a man-made distance but a complicated distance measured by a classifier network is used in GAN. This makes GAN's generated images more perceptual. 
It is expected that \textit{the generated images become much more perceptual when the power of the complicated classifier and the SSIM metric are combined.} 
The loss in GAN is a game-theoretical min-max problem:
\begin{align}\label{equation_GAN_loss}
\min_G \max_D \mathcal{L} = \mathbb{E}_{\b{x} \sim p(x)} \big[\log D(\b{x})\big] + \mathbb{E}_{\b{z} \sim q(z)} \big[\log (1 - D(G(\b{z})))\big],
\end{align}
where $p$ and $q$ are the distributions of the data and the latent variable, respectively, and $D$ and $G$ are the discriminator (classifier) and generator, respectively. The probability of data coming from real distribution is denoted by $D(.)$. The first term in Eq. (\ref{equation_GAN_loss}) is the log-likelihood of real data and the second term measures how different the generated and real data are. We can replace the second term by Eq. (\ref{equation_SSIM_distance}) or (\ref{equation_SSIM_distance2}), i.e., $||\b{x} - G(\b{z})||_S^2$, which is minimized and maximized by the generator and discriminator, respectively. This will \textit{measure how perceptually different the generated and real data are}, resulting in perceptually better generated images because \textit{both the generator and discriminator become more powerful in terms of perceptual differences}.
Conditional GAN \cite{mirza2014conditional} can use the same idea in its loss to have better generated images. 
AAE \cite{makhzani2015adversarial}, which is trained in an adversarial way like GAN, can also use the SSIM distance in its loss as was explained. 
Note that paper \cite{kancharla2018improving} has used a similar technique in GAN but with multi-scale SSIM (MS-SSIM) \cite{wang2003multiscale} used in Eq. (\ref{equation_SSIM_distance}) (see also \cite{snell2017learning}). 
Other ideas are used in some papers such as \cite{wu2017srpgan} which utilizes the middle-layer features of network for perceptually better generated images in GAN. 
On the other hand, LSGAN \cite{mao2017least} uses $\ell_2$ norm or MSE in its objective function to be optimized by the discriminator and generator. However, MSE is not suitable for perceptual image assessment. We propose to use SSIM distance, i.e. Eq. (\ref{equation_SSIM_distance}) or (\ref{equation_SSIM_distance2}), in place of the MSE terms in the LSGAN objectives to have perceptually better generated images. 

\section{Conclusion}

We theoretically analyzed how to use SSIM in generative models and inferential autoencoders including GMMN, VAE, AAE, GAN, and LSGAN. 
The use of SSIM in these models can improve the perceptual quality of the generated images as these models use non-perceptual distance metrics in their loss functions.

\bibliographystyle{splncs}      % basic style, author-year citations
\bibliography{references.bib}            % name your BibTeX data base

\begin{thebibliography}{1}

\bibitem{sample-citation}
Last, F., Last2, F.:
\newblock Canadian {{AI}}.
\newblock Canadian Artificial Intelligence \textbf{31} (2018)  1--12

\end{thebibliography}


\begin{thebibliography}{10}

\bibitem{ng2002discriminative}
Ng, A.Y., Jordan, M.I.:
\newblock On discriminative vs. generative classifiers: A comparison of
  logistic regression and naive {Bayes}.
\newblock In: Advances in neural information processing systems. (2002)
  841--848

\bibitem{doersch2016tutorial}
Doersch, C.:
\newblock Tutorial on variational autoencoders.
\newblock arXiv preprint arXiv:1606.05908 (2016)

\bibitem{wang2009mean}
Wang, Z., Bovik, A.C.:
\newblock Mean squared error: Love it or leave it? a new look at signal
  fidelity measures.
\newblock IEEE signal processing magazine \textbf{26}(1) (2009)  98--117

\bibitem{li2015generative}
Li, Y., Swersky, K., Zemel, R.:
\newblock Generative moment matching networks.
\newblock In: International Conference on Machine Learning. (2015)  1718--1727

\bibitem{ren2016conditional}
Ren, Y., Zhu, J., Li, J., Luo, Y.:
\newblock Conditional generative moment-matching networks.
\newblock In: Advances in Neural Information Processing Systems. (2016)
  2928--2936

\bibitem{makhzani2015adversarial}
Makhzani, A., Shlens, J., Jaitly, N., Goodfellow, I., Frey, B.:
\newblock Adversarial autoencoders.
\newblock arXiv preprint arXiv:1511.05644 (2015)

\bibitem{goodfellow2014generative}
Goodfellow, I., Pouget-Abadie, J., Mirza, M., Xu, B., Warde-Farley, D., Ozair,
  S., Courville, A., Bengio, Y.:
\newblock Generative adversarial nets.
\newblock In: Advances in neural information processing systems. (2014)
  2672--2680

\bibitem{mirza2014conditional}
Mirza, M., Osindero, S.:
\newblock Conditional generative adversarial nets.
\newblock arXiv preprint arXiv:1411.1784 (2014)

\bibitem{mao2017least}
Mao, X., Li, Q., Xie, H., Lau, R.Y., Wang, Z., Paul~Smolley, S.:
\newblock Least squares generative adversarial networks.
\newblock In: Proceedings of the IEEE International Conference on Computer
  Vision. (2017)  2794--2802

\bibitem{wang2004image}
Wang, Z., Bovik, A.C., Sheikh, H.R., Simoncelli, E.P.:
\newblock Image quality assessment: from error visibility to structural
  similarity.
\newblock IEEE transactions on image processing \textbf{13}(4) (2004)  600--612

\bibitem{otero2014unconstrained}
Otero, D., Vrscay, E.R.:
\newblock Unconstrained structural similarity-based optimization.
\newblock In: International Conference Image Analysis and Recognition, Springer
  (2014)  167--176

\bibitem{brunet2012mathematical}
Brunet, D., Vrscay, E.R., Wang, Z.:
\newblock On the mathematical properties of the structural similarity index.
\newblock IEEE Transactions on Image Processing \textbf{21}(4) (2012)
  1488--1499

\bibitem{zhao2016loss}
Zhao, H., Gallo, O., Frosio, I., Kautz, J.:
\newblock Loss functions for image restoration with neural networks.
\newblock IEEE Transactions on computational imaging \textbf{3}(1) (2016)
  47--57

\bibitem{ghojogh2019image}
Ghojogh, B., Karray, F., Crowley, M.:
\newblock Image structure subspace learning using structural similarity index.
\newblock In: International Conference on Image Analysis and Recognition,
  Springer (2019)  33--44

\bibitem{hofmann2008kernel}
Hofmann, T., Sch{\"o}lkopf, B., Smola, A.J.:
\newblock Kernel methods in machine learning.
\newblock The annals of statistics (2008)  1171--1220

\bibitem{brunet2018optimizing}
Brunet, D., Channappayya, S.S., Wang, Z., Vrscay, E.R., Bovik, A.C.:
\newblock Optimizing image quality.
\newblock In: Handbook of Convex Optimization Methods in Imaging Science.
\newblock Springer (2018)  15--41

\bibitem{gretton2012kernel}
Gretton, A., Borgwardt, K.M., Rasch, M.J., Sch{\"o}lkopf, B., Smola, A.:
\newblock A kernel two-sample test.
\newblock Journal of Machine Learning Research \textbf{13}(Mar) (2012)
  723--773

\bibitem{steinwart2001influence}
Steinwart, I.:
\newblock On the influence of the kernel on the consistency of support vector
  machines.
\newblock Journal of machine learning research \textbf{2}(Nov) (2001)  67--93

\bibitem{de1959stone}
De~Branges, L.:
\newblock The {Stone}-{Weierstrass} theorem.
\newblock Proceedings of the American Mathematical Society \textbf{10}(5)
  (1959)  822--824

\bibitem{harman1976modern}
Harman, H.H.:
\newblock Modern factor analysis.
\newblock University of Chicago press (1976)

\bibitem{kingma2013auto}
Kingma, D.P., Welling, M.:
\newblock Auto-encoding variational {Bayes}.
\newblock arXiv preprint arXiv:1312.6114 (2013)

\bibitem{kancharla2018improving}
Kancharla, P., Channappayya, S.S.:
\newblock Improving the visual quality of generative adversarial network
  ({GAN})-generated images using the multi-scale structural similarity index.
\newblock In: 2018 25th IEEE International Conference on Image Processing
  (ICIP), IEEE (2018)  3908--3912

\bibitem{wang2003multiscale}
Wang, Z., Simoncelli, E.P., Bovik, A.C.:
\newblock Multiscale structural similarity for image quality assessment.
\newblock In: The Thrity-Seventh Asilomar Conference on Signals, Systems \&
  Computers, 2003. Volume~2., IEEE (2003)  1398--1402

\bibitem{snell2017learning}
Snell, J., Ridgeway, K., Liao, R., Roads, B.D., Mozer, M.C., Zemel, R.S.:
\newblock Learning to generate images with perceptual similarity metrics.
\newblock In: 2017 IEEE International Conference on Image Processing (ICIP),
  IEEE (2017)  4277--4281

\bibitem{wu2017srpgan}
Wu, B., Duan, H., Liu, Z., Sun, G.:
\newblock {SRPGAN}: perceptual generative adversarial network for single image
  super resolution.
\newblock arXiv preprint arXiv:1712.05927 (2017)

\end{thebibliography}

\end{document}